\PassOptionsToPackage{numbers,sort&compress}{natbib}
\PassOptionsToPackage{breaklinks=true,bookmarks=false}{hyperref}
\pdfoutput=1

\documentclass[11pt]{article}

\usepackage{times}
\usepackage{latexsym}
\usepackage[T1]{fontenc}
\usepackage[utf8]{inputenc}
\usepackage{microtype}
\usepackage{inconsolata}
\usepackage{graphicx}

\usepackage[final]{coling}

\usepackage{amsmath}
\usepackage{amssymb}
\usepackage[group-separator={,}]{siunitx}

\usepackage{booktabs}
\usepackage{multirow}
\usepackage{tabularx}

\usepackage{listings}
\usepackage{xcolor}
\usepackage[numbers,sort&compress]{natbib}

\usepackage{subcaption}

\usepackage{dirtytalk}

\usepackage{fancyhdr}

\fancypagestyle{secondstyle}
{
   \fancyhf{}
   \fancyfoot[C]{{\raisebox{4ex}{\thepage}}}
}

\fancypagestyle{firststyle}
{
   \fancyhf{}
   \fancyfoot[C]{{\raisebox{4ex}{\large \thepage}}\\{{\raisebox{0pt}{\small 
\textit{Proceedings of the 31st International Conference on Computational Linguistics}, pages 31–35}}\\{\raisebox{3ex}{\small 
 January 19–24, 2025. ©COLING 2025}
}}}
}

\usepackage[breaklinks=true,bookmarks=false]{hyperref}

\begin{document}

\setcounter{page}{31}
\thispagestyle{firststyle}

\title{A Hybrid Approach to Information Retrieval and Answer Generation for Regulatory Texts}

\author{Jhon Rayo \\ 
  Universidad de los Andes \\
  Bogotá, Colombia \\
  \texttt{\small j.rayom@uniandes.edu.co} \\\And
  Raúl de la Rosa \\
  Universidad de los Andes \\
  Bogotá, Colombia \\
  \texttt{\small c.delarosap@uniandes.edu.co} \\\And
  Mario Garrido \\
  Universidad de los Andes \\
  Bogotá, Colombia \\
  \texttt{\small m.garrido10@uniandes.edu.co}}

\maketitle

\begin{abstract}
  Regulatory texts are inherently long and complex, presenting significant challenges for information retrieval systems in supporting regulatory officers with compliance tasks. This paper introduces a hybrid information retrieval system that combines lexical and semantic search techniques to extract relevant information from large regulatory corpora. The system integrates a fine-tuned sentence transformer model with the traditional BM25 algorithm to achieve both semantic precision and lexical coverage. To generate accurate and comprehensive responses, retrieved passages are synthesized using \textit{Large Language Models} (LLMs) within a \textit{Retrieval Augmented Generation} (RAG) framework. Experimental results demonstrate that the hybrid system significantly outperforms standalone lexical and semantic approaches, with notable improvements in Recall@10 and MAP@10. By openly sharing our fine-tuned model and methodology, we aim to advance the development of robust natural language processing tools for compliance-driven applications in regulatory domains.
\end{abstract}

\section{Introduction}

Information retrieval (IR) systems are concerned with efficiently querying large corpora to retrieve relevant results. Traditional systems, such as search engines, often depend on term-frequency statistical methods like \textit{tf-idf}, which measures the importance of a term in a document relative to its frequency in the corpus \citep{alma991005253134607681}. BM25 \citep{bm25}, a well-established ranking function, builds on similar principles to provide a scalable and effective retrieval framework. However, such methods are inherently limited when addressing complex domains like regulatory texts, where the semantics often outweigh simple term matching. \

\noindent Regulatory content is particularly challenging due to its specialized terminology and nuanced language. Synonyms, paraphrasing, and domain-specific jargon frequently obscure the relationship between queries and relevant documents, reducing the effectiveness of lexical retrieval methods. \

\noindent Semantic search addresses these limitations by using dense vector-based retrieval where we encode documents and queries as vectors, also known as \textit{embeddings}, capturing the semantic meaning of the text in a condensed high-dimensional space \citep{karpukhin2020densepassageretrievalopendomain}. This approach enables the system to measure similarity based on meaning rather than exact word matches, grouping related content together even with different terminology. \

\noindent Recent advances in pre-trained language models, like BERT \citep{bertpaper}, have introduced high-quality contextual \textit{embeddings} for words, sentences, and paragraphs which can be leveraged in semantic search applications.\

\noindent Despite these advances, building an effective IR system for regulatory texts poses unique challenges. Pre-trained language models are typically trained on general-purpose datasets and may lack the domain-specific knowledge required for accurate retrieval in specialized fields. Fortunately, various methods for transfer learning have demonstrated that these base models can be fine-tuned to close this gap \citep{houlsby2019parameterefficient}. \

\noindent In this paper, we present a hybrid information retrieval system that integrates both lexical and semantic approaches to address the limitations of traditional IR in the regulatory domain. Our method combines BM25 for lexical retrieval with a fine-tuned Sentence Transformer model \cite{reimers-2019-sentence-bert} to improve semantic matching. Additionally, we implement a Retrieval Augmented Generation (RAG) system \cite{ragpaper} that leverages the hybrid retriever to provide comprehensive and accurate answers to user queries using a Large Language Model (LLM). \

\noindent Through extensive experiments, we demonstrate that the hybrid retriever achieves superior performance compared to standalone lexical or semantic systems, as evidenced by improvements in Recall@10 and MAP@10. Furthermore, the RAG system effectively synthesizes retrieved content, delivering detailed responses that address the compliance requirements of regulatory questions. Our contributions aim to advance regulatory information retrieval and lay the foundation for more effective question-answering systems in specialized domains.

\section{Regulatory Information Retrieval}

The development of an effective information retrieval (IR) system for regulatory content requires addressing the unique challenges of compliance-related queries. These systems must return a set of ranked passages from the corpus that accurately address the compliance aspects of a given question. Previous work by \citet{gokhan2024regnlpactionfacilitatingcompliance} utilized BM25, a widely-used algorithm that ranks results based on query term frequency and other statistical features. While BM25 is effective for lexical retrieval, it struggles to capture semantic relationships, particularly in regulatory domains where terminology often varies for the same concepts. \

\noindent Our approach enhances BM25 by integrating a text embedding model, enabling semantic matching. This hybrid system identifies semantically relevant content that BM25 alone might overlook, offering a significant advantage in handling the complexities of regulatory language.

\subsection{Dataset}

The dataset used for this study, \textit{\textbf{ObliQA}}, consists of 27,869 regulatory questions extracted from 40 documents provided by Abu Dhabi Global Markets. This regulatory authority oversees financial services within the European Economic Area, making the dataset highly relevant for compliance-related tasks \cite{gokhan2024regnlpactionfacilitatingcompliance}. \

\noindent The dataset is divided into three subsets: training (22,295 questions), testing (2,786 questions), and validation (2,788 questions). Each question is paired with one or more passages that contain the relevant information needed to answer it. The data is stored in JSON format, where each entry includes the question, associated passages, and their metadata. An example is shown below. 
\newpage

\lstset{
    basicstyle=\ttfamily\small,
    numbers=left,
    numberstyle=\tiny,
    stepnumber=1,
    numbersep=5pt,
    backgroundcolor=\color{white},
    showstringspaces=false,
    keywordstyle=\color{blue},
    stringstyle=\color{green},
    commentstyle=\color{red},
    frame=single,
    breaklines=true,
    breakatwhitespace=true,
    tabsize=2
}

\begin{lstlisting}
{
  "QuestionID": "a10724b5-ad0e-4b69-8b5e-792aef214f86",
  "Question": "What are the two specific conditions related to the maturity of a financial instrument that would trigger a disclosure requirement?",
  "Passages": [
    {
      "DocumentID": 11,
      "PassageID": "7.3.4",
      "Passage": "Events that trigger a disclosure. For the purposes of Rules 7.3.2 and 7.3.3, a Person is taken to hold Financial ..."
    }
  ],
  "Group": 1
}
\end{lstlisting}

\subsection{Model Fine-tuning}

We fine-tuned the \textit{ \href{https://huggingface.co/BAAI/bge-small-en-v1.5}{BAAI/bge-small-en-v1.5}} \citep{bge_embedding}, a BERT-based model trained on general-purpose data. The fine-tuning process employed a loss function designed to maximize the similarity between questions and their associated passages. The architecture comprises a word embedding layer followed by pooling and normalization layers. To better capture semantic nuances in regulatory texts, we increased the embedding dimension from 384 to 512. \

\noindent Training was conducted on an NVIDIA A40 GPU with 24GB of memory using the \textit{SentenceTransformer} library \citep{reimers-2019-sentence-bert}. The model was trained over 10 epochs with a batch size of 64, using a learning rate of $2x10^{-4}$ to preserve the model's general-purpose knowledge while fine-tuning it for the domain. The \textit{MultipleNegativesRankingLoss} \citep{sbert_losses} loss function was employed, assuming all unpaired examples in the batch as negatives, which is particularly suited for scenarios with positive pairs only. \

\noindent Performance evaluation was conducted using the \textit{InformationRetrievalEvaluator} \citep{reimers2021informationretrievalevaluator} to compute metrics such as Recall@10, Precision@10, and MAP@10 during training. To further optimize the process, we employed warmup steps to gradually increase the learning rate, and Automatic Mixed Precision (AMP) \citep{zhao2021automatic} to reduce memory usage and enhance training speed. \

\noindent Table \ref{tab:model_results} summarizes the results, showing a significant performance improvement of the fine-tuned model over the base model in the regulatory domain. The fine-tuned model has been made available on \href{https://huggingface.co/raul-delarosa99/bge-small-en-v1.5-RIRAG_ObliQA}{Hugging Face Hub}, alongside the complete implementation in our \href{https://github.com/oyar99/IR-AG-REG/tree/main}{GitHub repository}.

\begin{table}[t]
  \centering
  \resizebox{\columnwidth}{!}{
  \begin{tabular}{lcc}
    \toprule
    \textbf{Model / Dataset} & \textbf{Recall@10} & \textbf{MAP@10} \\
    \midrule
    Base Model / Validation & 0.7135 & 0.5462 \\
    Base Model / Testing & 0.7017 & 0.5357 \\
    Custom Model / Validation & 0.8158 & 0.6315 \\
    Custom Model / Testing & \textbf{0.8111} & \textbf{0.6261}\\
    \bottomrule
  \end{tabular}
  }
  \caption{Performance comparison between the base model and the fine-tuned model.}
  \label{tab:model_results}
\end{table}

\subsection{Information Retrieval}

To enhance retrieval performance, we developed a data processing pipeline with the following steps:

\begin{enumerate} 
\item \textbf{Expand contractions}: Convert contractions (e.g., \textit{don't} to \textit{do not}) for consistency. 
\item \textbf{Normalization}: Lowercase text and remove non-alphanumeric characters using regular expressions. 
\item \textbf{Space removal}: Eliminate redundant spaces for uniformity. 
\item \textbf{Preserve legal format}: Retain special characters critical for legal documents.
\item \textbf{Stopwords}: Remove common words using \textit{nltk} and \textit{scikit-learn} sets.
\item \textbf{Stemming}: Apply the \textit{Snowball Stemmer} \citep{Porter2001SnowballAL} to reduce words to their root forms. 
\item \textbf{Tokenization}: Generate unigrams and bigrams to capture both individual terms and word combinations. 
\end{enumerate}

\noindent Using this pipeline, we implemented three retrieval approaches:
\begin{enumerate}
    \item BM25 (Baseline): Configured with $k=1.5$ and $b=0.75$.
    \item Semantic Retriever: Leveraged the fine-tuned model for semantic matches only.
    \item Hybrid System: Combined BM25 and the fine-tuned model, computing an aggregated score using Equation \ref{eq:score}: 
    \begin{equation}
    \begin{aligned}
      \text{Score} = \alpha &\cdot \text{Semantic Score} \\ 
      &+ (1 - \alpha) \cdot \text{Lexical Score}
    \label{eq:score}
    \end{aligned}
    \end{equation}
\end{enumerate}

\noindent We empirically set $\alpha = 0.65$ to give slightly higher weight to semantic matching while maintaining meaningful contribution from lexical search. This normalization step ensures that neither approach dominates the final ranking purely due to differences in score distributions.

\noindent Table \ref{tab:retrieval_results} compares the performance of these approaches. The hybrid system demonstrates the highest effectiveness, combining the strengths of lexical and semantic retrieval methods.

\begin{table}[t]
  \centering
  \resizebox{\columnwidth}{!}{
  \begin{tabular}{lcccc}
    \toprule
    \textbf{Model} & \textbf{Recall@10} & \textbf{MAP@10} & \textbf{Recall@20} & \textbf{MAP@20} \\
    \midrule
    BM25 (Baseline) & 0.7611 & 0.6237 & 0.8022 & 0.6274 \\
    BM25 (Custom) & 0.7791 & 0.6415 & 0.8204 & 0.6453 \\
    Semantic system & 0.8103 & 0.6286 & 0.8622 & 0.6334 \\
    Hybrid system & \textbf{0.8333} & \textbf{0.7016} & \textbf{0.8704} & \textbf{0.7053} \\
    \bottomrule
  \end{tabular}
  }
  \caption{Performance comparison between information retrieval systems.}
  \label{tab:retrieval_results}
\end{table}

\section{Answer Generation}

\textit{Retrieval Augmented Generation} (RAG) is a cutting-edge technique that enhances \textit{Large Language Models} (LLMs) by integrating external retrieval capabilities, enabling them to generate responses based on information they were not explicitly trained on \citep{ragpaper}. This approach has emerged as a powerful tool in open-domain question-answering applications, combining retrieval-based and generation-based methods to improve answer relevance and quality \citep{siriwardhana2023improving}. \

\noindent In our system, RAG is used to answer regulatory questions by leveraging the hybrid information retrieval system described earlier. The retrieved passages provide the contextual foundation for generating answers that address compliance-related aspects comprehensively and accurately. \

\noindent Given a regulatory question, similar to the approach followed in \cite{gokhan2024regnlpactionfacilitatingcompliance}, the system retrieves up to 10 relevant passages from the corpus. To ensure high-quality input for the answer generation process, only passages with a relevance score of at least $0.72$ are considered. Additionally, passage processing is terminated when the relevance score drops by more than $0.1$ from the previous passage, maintaining the relevance and coherence of the input data. \

\noindent These selected passages are fed into an LLM to synthesize a concise and coherent answer. For this task, we experimented with three different models: \textit{GPT 3.5 Turbo} and \textit{GPT-4o Mini} through Azure OpenAI batch deployment, and \textit{Llama 3.1} using Groq's API. When evaluated on our test dataset, \textit{GPT 3.5 Turbo} achieved the highest RePASs score of 0.57, significantly outperforming both \textit{GPT-4o Mini} (0.44) and \textit{Llama 3.1} (0.37), leading to its selection as our primary model. We designed the system prompt to guide response generation in the regulatory domain, emphasizing accuracy, completeness, and alignment with the provided passages. The prompt reads:

\begin{quote}
    \small\say{\textit{As a regulatory compliance assistant. Provide a **complete**, **coherent**, and **correct** response to the given question by synthesizing the information from the provided passages. Your answer should **fully integrate all relevant obligations, practices, and insights**, and directly address the question. The passages are presented in order of relevance, so **prioritize the information accordingly** and ensure consistency in your response, avoiding any contradictions. Additionally, reference **specific regulations and key compliance requirements** outlined in the regulatory content to support your answer. **Do not use any extraneous or external knowledge** outside of the provided passages when crafting your response.}}
\end{quote}

\noindent We selected the top 3 answers with the highest RePASs scores to enhance the prompt using few-shot techniques, aiming to improve its performance. Below is a demonstration of how we used this prompting method.

\begin{quote}
    \small\say{\textit{
    Question: What percentage of the Insurer's Net Written Premium is used to determine the non-proportional reinsurance element?
    Passage: The non proportional reinsurance element is calculated as  of the Insurer's Net Written Premium
    Your response should read:
    The non-proportional reinsurance element is determined by calculating 52 percent of the Insurer's Net Written Premium.}}
\end{quote}

\noindent \textit{Regulatory Passage Answer Stability Score} (RePASs), introduced by \citet{gokhan2024regnlpactionfacilitatingcompliance} assesses the stability and accuracy of generated answers across three key dimensions:
\begin{enumerate}
    \item Entailment Score ($E_s$): Measures the extent to which each sentence in the generated answer is supported by sentences in the retrieved passages.
    \item Contradiction Score ($C_s$): Evaluates whether any sentence in the generated answer contradicts the information in the retrieved passages.
    \item Obligation Coverage Score ($OC_s$): Checks if the generated answer covers all obligations present in the retrieved passages.
\end{enumerate}

\noindent The composite RePASs score is derived from these metrics, offering a holistic measure of the system's answer quality. Table \ref{tab:results} summarizes the evaluation results, comparing our approach to the baseline. \

\noindent Table \ref{tab:results} shows that while our system achieves moderate improvements in obligation coverage ($OC_s$) and slightly better contradiction handling ($C_s$), its entailment score ($E_s$) reveals areas for further optimization. The hybrid retrieval system enhances answer relevance by incorporating semantic and lexical matches, but the synthesis process using \textit{GPT 3.5 Turbo} shows reduced performance in capturing the degree to which generated answers are supported by the retrieved passages, as evidenced by the lower entailment score.

\begin{table}[t]
  \centering
  \resizebox{\columnwidth}{!}{
  \begin{tabular}{lcccc}
    \toprule
    \textbf{System} & \textbf{Es} & \textbf{Cs} & \textbf{OCs} & \textbf{RePASs} \\
    \midrule
    Baseline & 0.78 & 0.24 & 0.20 & 0.58 \\
    Hybrid retriever + GPT-4o Mini & 0.38 & 0.23 & 0.17 & 0.44 \\
    Hybrid retriever + Llama 3.1 & 0.34 & 0.45 & 0.22 & 0.37 \\
    Hybrid retriever + GPT 3.5 Turbo & \textbf{0.58} & \textbf{0.21} & \textbf{0.33} & \textbf{0.57} \\
    \bottomrule
  \end{tabular}
  }
  \caption{Performance comparison of answer generation systems using RePASs metrics.}
  \label{tab:results}
\end{table}

\section{Conclusion}

This work tackles the significant challenges of retrieving and synthesizing information from complex regulatory texts by demonstrating the effectiveness of hybrid approaches that integrate lexical and semantic retrieval methods. Our results show the importance of combining classical algorithms, such as BM25, with embedding-based models to address the nuanced language and diverse terminologies inherent in regulatory domains. The hybrid system consistently outperforms standalone lexical or semantic approaches, achieving notable improvements in metrics like Recall@10 and MAP@10. \

\noindent We further demonstrate the potential of LLMs to synthesize concise and comprehensive answers. These models effectively utilize the structured information retrieved by the hybrid system to address regulatory queries with improved coherence and relevance. However, the evaluation using RePASs reveals opportunities for refinement, particularly in improving entailment metrics. \

\noindent Future directions include fine-tuning LLMs on domain-specific corpora to enhance alignment with regulatory contexts, optimizing retrieval thresholds for better semantic coverage, and exploring advanced scoring mechanisms to balance precision and recall.

\section*{Acknowledgments}
This work was supported by the NLP Group at Universidad de los Andes. We thank Abu Dhabi Global Markets for providing access to their regulatory documents. Special thanks to our dedicated professor Rubén Francisco Manrique.

{\small
\bibliography{custom}  
}

\appendix
\end{document}